\documentclass{article}
\usepackage{float}
\usepackage{placeins}
\usepackage[preprint]{neurips_2026}
\makeatletter
\renewcommand{\@noticestring}{Accepted to WM@Booth 2026: Workshop on World Models, hosted by Chicago Booth.}
\makeatother

\usepackage[utf8]{inputenc}
\usepackage[T1]{fontenc}
\usepackage{hyperref}
\usepackage{url}
\usepackage{booktabs}
\usepackage{amsfonts}
\usepackage{nicefrac}
\usepackage{microtype}
\usepackage[table]{xcolor}
\usepackage{multirow}
\usepackage{makecell}
\usepackage{caption}
\usepackage{array}
\usepackage{amssymb}
\usepackage{amsmath}
\usepackage{graphicx}
\usepackage{subcaption}

\definecolor{stagedrow}{HTML}{EAF7EA}
\definecolor{bestflat}{HTML}{E8F1FF}

\newcommand{\meanstd}[2]{%
  \makebox[2.2em][r]{#1}\,\ensuremath{\pm}\,\makebox[1.7em][l]{#2}%
}

\hypersetup{hidelinks}

\title{Mind the Gap: Promises and Pitfalls of Hierarchical Planning in LeWorldModel}

\author{%
  Niccol\'o Caselli$^{*}$\\
  University of Amsterdam\\
  \And
  Francesco Massafra$^{*}$\\
  University of Amsterdam\\
  \And
  Samuele Punzo$^{*}$\\
  University of Amsterdam\\
  \AND
  Salvatore Lo Sardo$^{*}$\\
  University of Amsterdam\\
  \And
  Ippokratis Pantelidis$^{*}$\\
  University of Amsterdam\\
  \And
  Sathya Kamesh Bhethanabhotla\\
  University of Amsterdam\\
  \AND
  $^{*}$Equal contribution.
}

\begin{document}

\maketitle

\begin{abstract}Effective long-horizon planning using latent world models remains a formidable challenge, and temporal abstraction is a promising direction for addressing it. We investigate whether this abstraction can improve LeWorldModel on long-horizon, goal-conditioned control tasks. To this end, we introduce Hi-LeWM, an extension that freezes the pretrained low-level LeWM and adds a high-level predictor that plans over latent subgoals. We evaluate Hi-LeWM on PushT and OGBCube across increasing goal offsets; our results suggest that hierarchy does not automatically improve performance. At short horizons, the best configuration benefits from a one-step high-level plan, while longer horizons reveal a mismatch between the learned macro-action space and the inference-time distribution, which we trace to the quality of CEM-selected subgoals as the main bottleneck. This is because unconstrained search can select latent macro-actions that appear favorable under the learned model but yield poor control targets. We find that constraining the search around macro-actions encoded from training trajectories, combined with appropriate subgoal execution timing, recovers effective hierarchical regimes, improving over flat LeWM by $+11.3\%$ at medium-range horizons and $+14.7\%$ at the longest PushT horizon. Overall, temporal abstraction can benefit long-horizon planning in LeWM, but only when paired with a constrained high-level search space.\end{abstract}

\section{Introduction}
\label{sec:introduction}

Latent world models~\citep{worldmodel} enable embodied agents to plan in compact representation spaces rather than pixels~\citep{hafner2020dreamer}. LeWorldModel (LeWM)~\citep{maes2026leworldmodel} follows this paradigm with a compact JEPA-style latent dynamics model~\citep{JEPA} and CEM planning~\citep{CEM} toward goal observations. Its simplicity is appealing: it avoids pixel prediction and keeps planning directly tied to latent control. However, long-horizon control remains difficult. Model errors compound over imagined rollouts, while optimizing sequences of primitive actions becomes increasingly challenging as the planning horizon grows. In this regard, LeWM remains a flat planner: it operates directly over primitive actions at a single temporal scale, which is effective for short-horizon goals but less well suited to longer-horizon tasks.

Temporal hierarchies are a natural candidate for addressing this limitation~\citep{gumbsch2024thick,hafner2022director}. We therefore ask whether temporal abstraction can be added to LeWM as a minimal extension, without departing from its simple JEPA-style latent prediction and CEM-based planning framework.

Rather than designing a new hierarchy, we ask whether the waypoint-based latent-planning interface from recent hierarchical world-model work transfers to LeWM as a lightweight retrofit. This setting is deliberately restrictive: the low-level LeWM planner remains fixed, and the hierarchy must succeed by adding only a high-level macro-action model and planner.

To study this, we introduce Hi-LeWM, which keeps the pretrained LeWM low-level model fixed and adds a high-level latent dynamics model over macro-actions. The high level proposes intermediate latent subgoals, while the original low-level planner acts toward them.

We evaluate Hi-LeWM on two goal-conditioned control benchmarks, PushT \citep{Chi2023DiffusionPV} and OGBCube \citep{park2025ogbench}, across increasing goal offsets. Our results show that, for LeWorldModel, simply adding a temporal abstraction layer on top of the frozen flat planner is not sufficient to improve long-horizon control. The straightforward Hi-LeWM planner often underperforms flat LeWM, so we analyze why the added hierarchy can become a liability rather than an advantage. Oracle intermediate subgoals are often executable, whereas generated subgoals are less reliable, temporally misaligned, and sensitive to the high-level search space. Constraining high-level search with data-supported macro-actions and choosing an appropriate execution mode recovers useful behavior in some settings, but does not provide a general solution.

Taken together, this study contributes a lightweight hierarchical extension of LeWM, an empirical evaluation of when this added temporal abstraction fails to improve the frozen flat planner, and an explanatory analysis showing that planner-induced support mismatch, support-constrained search, and execution mode determine when the hierarchy becomes useful in our experiments.

\subsection{Related Work}

Planning in learned latent spaces has been a central theme in model-based control from pixels. Early world-model planners such as PlaNet~\citep{hafner2019planet} showed that compact latent dynamics can support online planning directly from image observations, and later methods such as TD-MPC~\citep{hansen2022tdmpc}, DINO-WM~\citep{dinowm}, and LeWM~\citep{maes2026leworldmodel} further emphasized task-oriented latent prediction and test-time trajectory optimization. Our setting is most aligned with this line of work: we study goal-conditioned latent planning under a fixed planning budget rather than learning a separate policy over abstract actions.

A second line of work addresses long-horizon control through temporal abstraction. THICK~\citep{gumbsch2024thick} learns hierarchical world models with adaptive temporal abstractions, while recent hierarchical latent-planning methods explicitly plan across multiple temporal scales and report gains on non-greedy long-horizon tasks~\citep{zhang2026hierarchical, hafner2022director}. These results motivate hierarchy, but they also typically introduce a different model class, training pipeline, or planning interface from the base flat planner. Our goal is narrower: we isolate what happens when hierarchy is added as a minimal retrofit on top of a frozen LeWM backbone. Recent work also suggests that long-horizon failure is not only a modeling problem but can also be a search problem: IMWM~\citep{gao2026imwm} shows that finite-budget sample-based planning can remain a bottleneck even with a strong world model. Our findings are consistent with that view, and further localize the issue in Hi-LeWM to planner-induced support mismatch in the high-level search space.

\section{Method: Hi-LeWM and High-Level Planning}
\label{sec:methods}

\paragraph{LeWorldModel}
     LeWM~\citep{maes2026leworldmodel} performs goal-conditioned control by planning in a learned latent space. An encoder maps the observation and goal image to latent representations $z_t=f_\theta(o_t)$ and $z_g=f_\theta(o_g)$. Given a candidate action sequence, the frozen low-level action-conditioned predictor $p_{\mathrm{lo}}$ rolls the current latent state forward, producing $\hat{z}_{t+H}=p_{\mathrm{lo}}(z_t,a_{t:t+H-1})$. Planning then selects the action sequence whose predicted terminal latent is closest to the goal latent:
\[
  a^*_{t:t+H-1}
  =
  \underset{a_{t:t+H-1}}{\text{argmin}}
  \left\|
  \hat{z}_{t+H} - z_g
  \right\|_2^2 .
\]
The optimization is carried out with CEM in a receding-horizon loop, executing only a short prefix of each planned sequence before replanning. LeWM is therefore a flat planner: it searches directly over low-level actions and scores them by predicted progress toward the goal in latent space.

\paragraph{Hierarchical LeWorldModel (Hi-LeWM)}
Hi-LeWM intentionally follows the waypoint-based hierarchical latent-planning interface of recent work~\citep{zhang2026hierarchical}: a high-level model predicts latent waypoints by optimizing over learned macro-actions, and a low-level planner acts toward the selected waypoint. Following this setup, we define latent waypoints as encoded observations sampled at temporally separated states along a trajectory. The pretrained low-level LeWM components are kept frozen, while we learn two additional high-level modules: a latent macro-action encoder $g$ modeled as a transformer and a high-level predictor $p_{\mathrm{hi}}$ modeled as a transformer with adaptive layer normalization~\citep{DBLP:journals/corr/HuangB17}. Given a temporally extended primitive action chunk $a_{t:t+k-1}$ between two consecutive waypoints, the encoder produces a latent macro-action $\ell_t = g(a_{t:t+k-1})$, which summarizes the behavior executed between them. The high-level predictor then models abstract latent transitions as $\hat z_{t+k}=p_{\mathrm{hi}}(z_t,\ell_t)$,
where $z_t=f_\theta(o_t)$ is the latent embedding of the current observation and $k$ denotes the temporal gap between waypoints. In this way, Hi-LeWM augments LeWM with an additional dynamics model operating at a coarser temporal scale while preserving the original latent interface.

At test time, Hi-LeWM plans at two temporal scales. A high-level CEM planner searches over a sequence of latent macro-actions $\ell_{1:h_h}$, where $h_h$ is the high-level planning horizon. Recursively applying $p_{\mathrm{hi}}$ yields a terminal high-level prediction $\hat z^{\mathrm{hi}}_{h_h}$, and CEM selects the sequence whose terminal prediction is closest to the goal latent:
\[
\ell^*_{1:h_h}
=
\underset{\ell_{1:h_h}}{\text{argmin }}
\left\|
\hat z^{\mathrm{hi}}_{h_h}-z_g
\right\|_2^2 .
\]
The first predicted subgoal is then passed to the frozen low-level LeWM planner, which optimizes a short primitive action sequence toward that subgoal. Only a portion of the low-level plan is executed before replanning, following standard receding-horizon MPC. Additional details on training, waypoint construction, and the full objective are provided in Appendix~\ref{app:Hi-LeWM-details}.

\section{Benchmark Results}

We evaluate primarily on \texttt{PushT-v1}, where goal-conditioned episodes can be constructed with different temporal offsets between the initial observation and the goal. This lets us study the same control problem across increasingly long planning horizons. \texttt{OGBCube-v0} results are reported in Appendix~\ref{app:cube-results}; because its flat baseline remains comparatively stable as $d$ increases, Cube is less diagnostic of long-horizon planning issues. In both tasks, the agent observes image states and must reach a future goal observation. Experiments use the stable-worldmodel library~\citep{maes_lld2026swm}. Following prior hierarchical latent world-model evaluation practice~\citep{zhang2026hierarchical}, we vary the temporal distance between the start state and the goal using offsets $d \in \{25,50,75\}$ environment steps. For each model and offset, we sweep planning hyperparameters and report the best success rate across the sweep, averaged over three random seeds. Full planning hyperparameters are reported in Appendices~\ref{app:Hi-LeWM-planning-configs} and~\ref{app:Hi-LeWM-cem}.

\subsection{Preliminary Results: Naive High-Level Planning Falls Short}

\paragraph{Baseline Reproduction.}
We first reproduce the original LeWM baseline under our goal-conditioned evaluation protocol. This establishes the performance of the flat planner before introducing temporal hierarchy. On PushT, the flat LeWM baseline reaches $94.0\%$ at $d{=}25$, $52.7\%$ at $d{=}50$, and $18.0\%$ at $d{=}75$. This pattern gives a clear reference point for the rest of the paper: hierarchy should not be expected to help at short horizons where flat planning is already strong, but may become useful once the goal offset exceeds the reliable range of flat CEM planning. The corresponding flat planning configurations are reported in Appendix~\ref{app:Hi-LeWM-planning-configs}.

\paragraph{Hi-LeWM Benchmark Results.}
The naive hierarchical extension does not solve this degradation. With standard high-level CEM, Hi-LeWM reaches $89.3\%$, $38.7\%$, and $15.3\%$ at $d\in\{25,50,75\}$, respectively. This is disappointing in precisely the regime where hierarchy is meant to help: at $d{=}50$, naive Hi-LeWM is below the flat baseline by $14.0\%$, and at $d{=}75$ it remains $2.7\%$ below the already weak flat planner. The $d{=}25$ result is also informative. The best sweep uses a high-level horizon of $h_h{=}1$, so the strongest short-horizon hierarchical configuration is effectively a shallow one-step subgoal planner rather than a useful temporal abstraction. These results identify a transfer gap: a waypoint-based macro-action planner motivated by prior hierarchical latent-planning work is not sufficient, by itself, to improve a compact LeWM planner with frozen low-level components. These preliminary results motivate the central question of the paper: why does adding a high-level model and high-level CEM make planning worse before it makes it better? We answer this with diagnostics in Section~\ref{sec:failure-analysis}, and only then introduce the planning-side fixes in Section~\ref{sec:planning-mitigations}.

\subsection{Failure Analysis: Subgoal Quality and Planner Exploitation}
\label{sec:failure-analysis}
Benchmark success rates do not directly identify which component of the Hi-LeWM hierarchy fails. We therefore run two PushT diagnostics at $d{=}50$: an acting decomposition that replaces learned subgoals with oracle future latents, and an offline high-level forecast diagnostic that compares true and CEM-selected macro-actions.
To visualize these diagnostics, we use the lightweight transformer decoder probe described in Appendix~\ref{app:decoder_probe}, which maps latent states back to pixel space. The probe is used only for analysis and figure generation, not for policy execution.

\begin{figure}[htbp]
  \centering
  \includegraphics[width=0.86\linewidth]{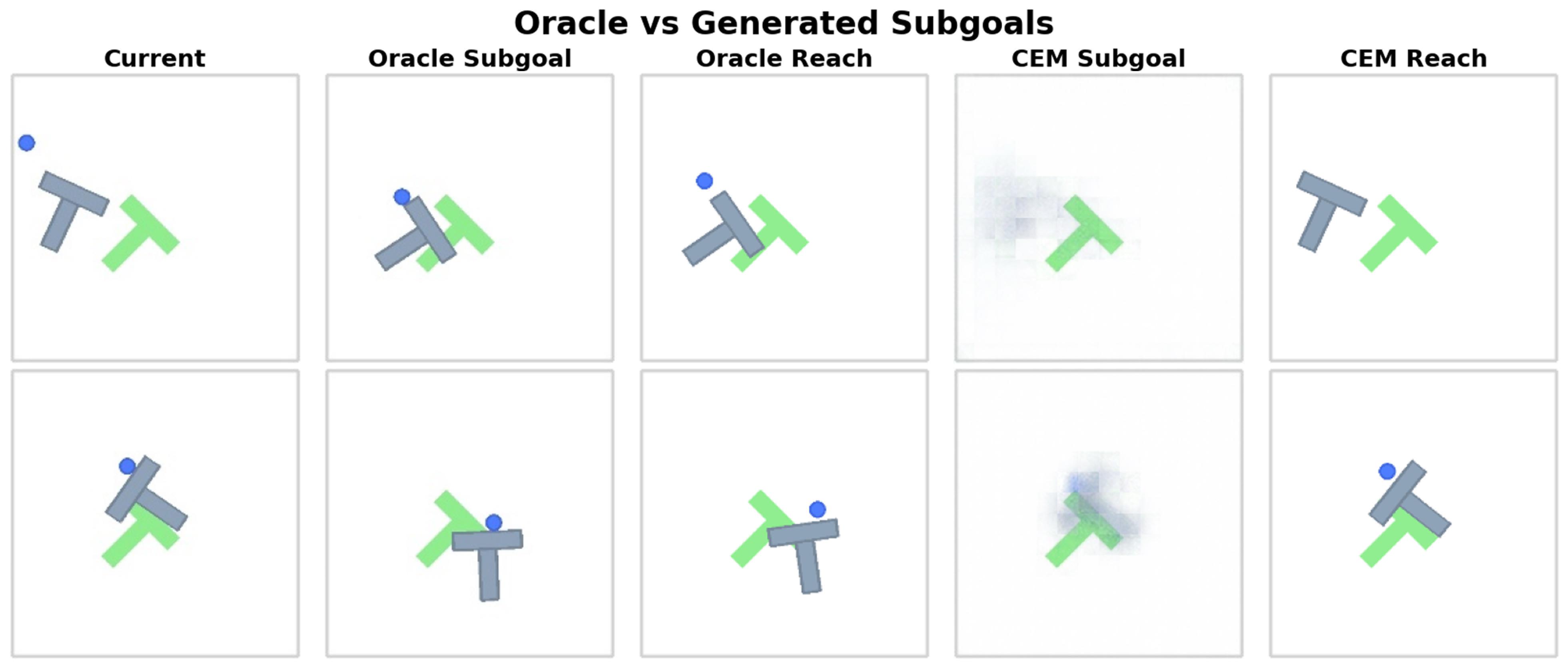}
  \caption{Qualitative PushT subgoal diagnostic. Columns show the current observation, an oracle future-observation target, the state reached when acting toward it, the CEM-generated subgoal, and the state reached from that target. Only the CEM Subgoal column is decoded with the probe.}
  \label{fig:oracle_vs_generated}
\end{figure}

\paragraph{Acting Decomposition.}
We first test whether the frozen low-level planner can execute valid intermediate targets. Oracle-subgoal acting replaces the high-level planner with true future latents from the dataset, while generated-subgoal acting uses learned high-level targets. On PushT at $d{=}50$, oracle-subgoal acting reaches $73.3 \pm 4.2\%$ success, indicating that the hierarchical controller can work when intermediate latent targets are well aligned with the task trajectory. Figure~\ref{fig:oracle_vs_generated} shows that oracle targets are executed more reliably, whereas generated targets are diffuse, poorly localized, and inconsistent with reachable intermediate configurations. This suggests that the main failure is not low-level executability alone, but the quality of high-level subgoal generation. 

\paragraph{High-Level Forecast Diagnostic.}
We next analyze the high-level model in isolation, before executing its plans. Table~\ref{tab:CEMDiagnostic} compares three rollout modes. In the teacher-forced setting, each prediction is conditioned on the true previous latent state. In open-loop true, the model is rolled forward autoregressively using true macro-actions from the dataset. In open-loop CEM, the same autoregressive rollout is driven by macro-actions selected by high-level CEM. True open-loop prediction shows the expected error increase across stages, but the errors remain small. By contrast, CEM-selected macro-actions produce a poor first-stage prediction while achieving the lowest second-stage error. This indicates that CEM can exploit the terminal latent objective by selecting macro-actions that look good at the final step but do not yield useful intermediate subgoals for control.

\begin{table}[htbp]
  \centering
  \small
  \setlength{\tabcolsep}{5.5pt}
  \renewcommand{\arraystretch}{1.08}
  \captionsetup{skip=4pt}
  \begin{tabular}{lccc}
    \toprule
    & Teacher-forced & Open-loop true & Open-loop CEM \\
    \midrule
    Stage 1 MSE $\downarrow$
    & 0.081
    & 0.081
    & \textbf{0.347} \\
    Stage 2 MSE $\downarrow$
    & 0.104
    & 0.216
    & \textbf{0.011} \\
    \bottomrule
  \end{tabular}
  \caption{High-level forecast diagnostic comparing teacher-forced, open-loop true, and open-loop CEM-selected rollouts. Stage~1 and Stage~2 denote the first and second high-level subgoal predictions in the two-stage rollout.}
  \label{tab:CEMDiagnostic}

\end{table}

\subsection{Planning-Side Mitigations}
\label{sec:planning-mitigations}

The diagnostics above suggest a targeted mitigation. The problem is not only that the model must predict farther into the future; it is that high-level planning creates a latent macro-action search space that the optimizer can exploit. During training, macro-actions are observed only as encoder outputs of real action chunks. At test time, however, standard CEM searches continuous latent macro-actions freely, including regions that may have low predicted terminal cost but poor intermediate control meaning. The first mitigation is therefore to constrain high-level CEM to remain close to data-supported macro-actions.

\begin{figure}[htbp]
    \centering
    \includegraphics[width=0.86\linewidth]{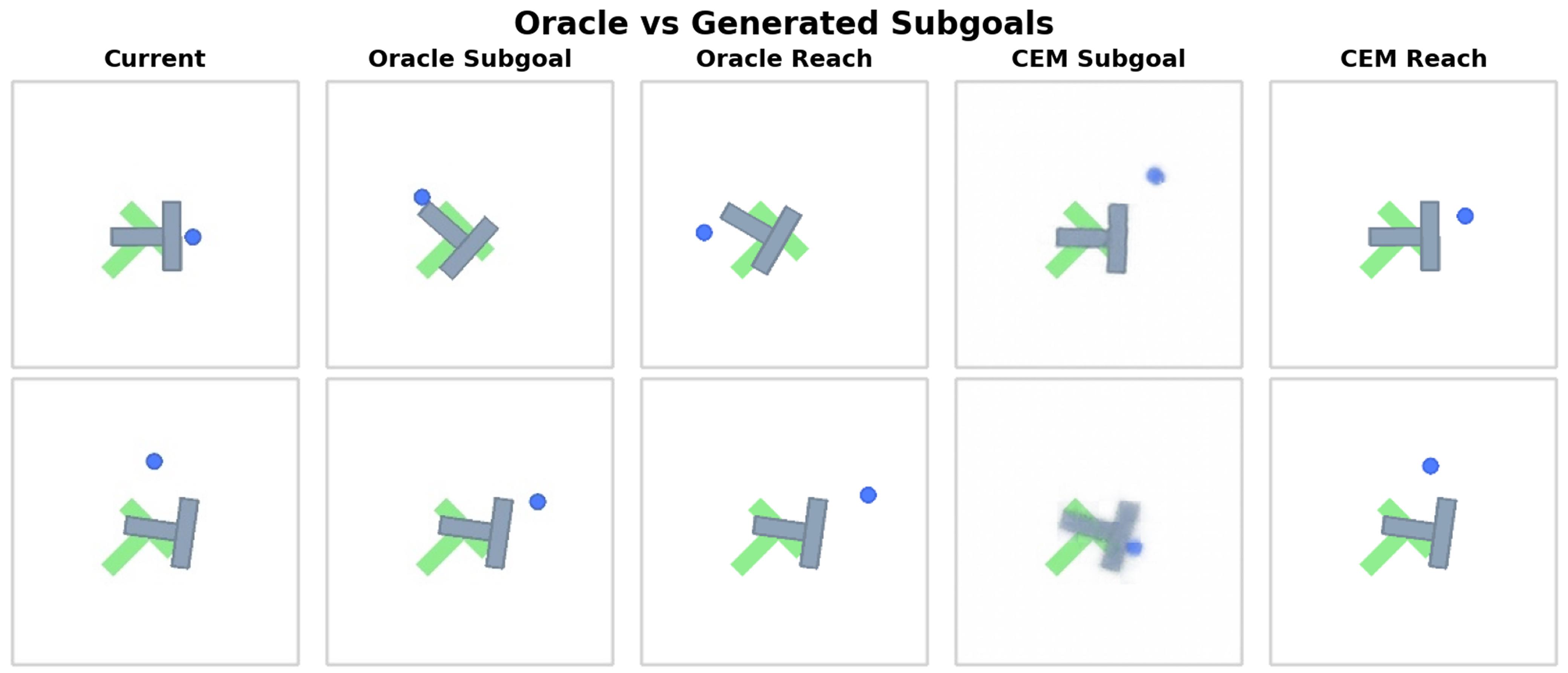}
    \caption{Qualitative PushT diagnostic under empirical-macro CEM, using the same column layout as Figure~\ref{fig:oracle_vs_generated}. Compared with unconstrained CEM, the generated targets are visually sharper and the reached states better match the intended intermediate configurations.}
    \label{fig:constrained_cem_subgoals}
\end{figure}

We implement this as empirical-macro CEM. Instead of sampling high-level candidate sequences directly from an unconstrained Gaussian, we build a bank of macro-action sequences encoded from training trajectories and sample candidates as local residual perturbations around these empirical anchors. The planner still performs CEM optimization, but the search distribution is biased toward macro-actions that could have been produced by the training action stream and macro-action encoder. This turns the diagnostic finding into an inference-time support constraint. Consistent with this mechanism, the empirical-macro planner produces sharper CEM subgoals that are more aligned with reachable PushT configurations (Figure~\ref{fig:constrained_cem_subgoals}). We also explored alternative constraints, including waypoint variants, reduced continuous latent dimensions, and vector-quantized macro-actions; these are summarized in Appendix~\ref{ssec:VQ-vae} and \autoref{tab:Hi-LeWM-representation-results}. We focus on empirical-macro CEM in the main analysis because it directly constrains high-level search at inference time, whereas the representation variants require retraining and are better treated as exploratory architecture choices. Full empirical-macro CEM details are given in Appendix~\ref{app:empirical_macro_cem}.

We hypothesize that staged execution may be most useful at intermediate horizons, where CEM can find a coherent waypoint sequence but repeated high-level replanning may disrupt its execution. We therefore test staged execution alongside online replanning. In the standard online setting, Hi-LeWM periodically replans the high level and passes only the first predicted waypoint to the low-level controller. In staged execution, high-level CEM runs once at the beginning of the episode, stores an ordered list of latent targets, and advances the active target after a fixed stage duration. The low-level planner still replans in closed loop toward the current target. This preserves a generated temporal scaffold, but cannot correct an inaccurate high-level rollout online.

\begin{figure}[htbp]
  \centering
  \includegraphics[width=0.75\linewidth]{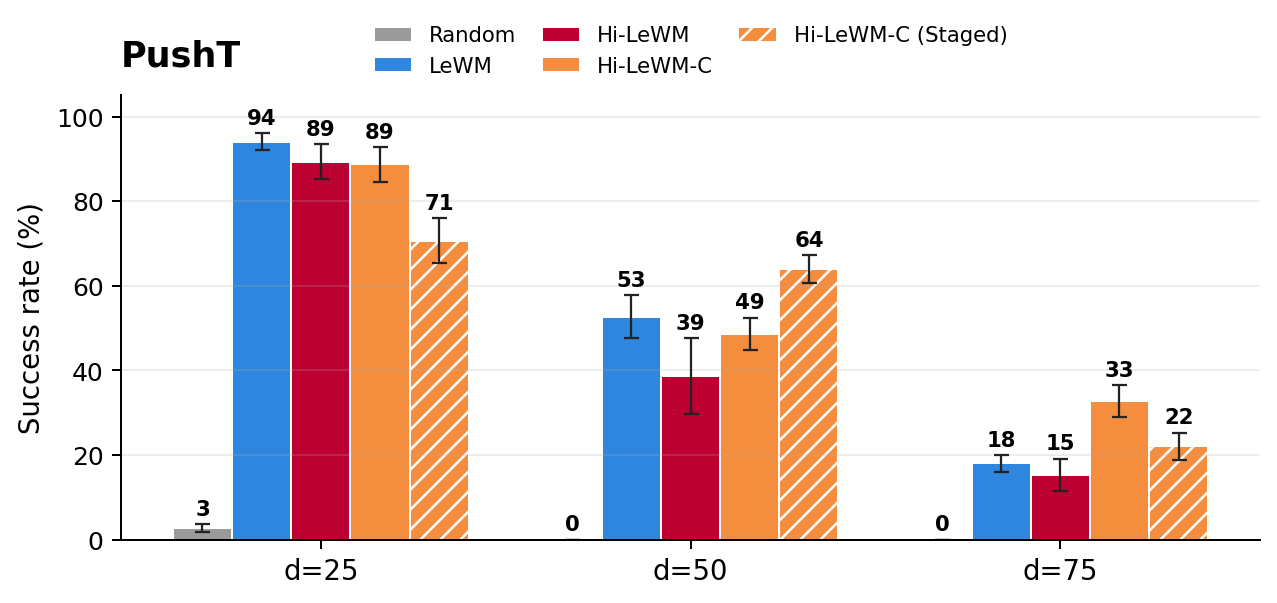}
  \caption{PushT success rates across goal offsets, averaged over three random seeds. For each model and goal offset, we report the best configuration found by the corresponding hyperparameter sweep. The full sweep results and selected planning configurations are reported in Appendix~\ref{app:Hi-LeWM-planning-configs}.}
  \label{fig:pusht_results}
\end{figure}

Figure~\ref{fig:pusht_results} summarizes the effect of these planning-side mitigations. The best non-staged empirical-macro configuration selected by the sweep improves the online hierarchical planner from $38.7\%$ to $48.7\%$ at $d{=}50$ and from $15.3\%$ to $32.7\%$ at $d{=}75$. Adding staged execution gives the clearest win at $d{=}50$, reaching $64.0\%$ and improving by $11.3\%$ over flat LeWM. At $d{=}75$, however, staged execution falls to $22.0\%$, below online Hi-LeWM-C. Thus the fix has a natural limit: support-constrained search helps, but committing to a long staged sequence helps only when the temporal scale is appropriate.

\FloatBarrier

\section{Discussion and Conclusion}
Our results show that temporal abstraction can help a compact LeWM planner with frozen low-level components, but only when the high-level search space remains compatible with the low-level controller. Hi-LeWM deliberately keeps the pretrained LeWM encoder, low-level predictor $p_{\mathrm{lo}}$, and low-level action encoder fixed, and adds only a high-level predictor $p_{\mathrm{hi}}$ and macro-action encoder $g$. This isolates whether hierarchy can be added as a lightweight addition rather than by redesigning the full controller. The resulting picture is conditional. At $d{=}25$, flat LeWM is already near saturated and the best hierarchical sweep uses $h_h{=}1$, so hierarchy provides no meaningful advantage. At $d{=}50$, empirical-macro CEM with staged execution reaches $64.0\%$ success, improving over flat LeWM by $+11.3\%$. At $d{=}75$, staged execution is not effective, but online constrained replanning reaches $32.7\%$, improving over flat LeWM by $+14.7\%$.

The main failure mode is therefore not low-level executability alone, nor temporal abstraction alone. Unconstrained high-level CEM can search macro-actions outside the support induced by training trajectories, producing subgoals that appear favorable under the learned terminal objective but are poor targets for control. Empirical-macro CEM mitigates this by turning high-level planning into local search around data-supported macro-action sequences. Its benefit, however, depends on the execution mode and temporal scale: support constraints improve the search distribution, while staged execution helps only when the generated sequence remains reliable over the committed interval.

\paragraph{Limitations and Future Work.}
This study extends LeWM into a hierarchical regime and, in doing so, uncovers structural tensions that are not visible from flat planning alone. Our results should be interpreted as an analysis of hierarchy in LeWM rather than a complete evaluation of hierarchical world-model design. We primarily use PushT because varying the goal offset $d$ gives a controlled way to study longer-horizon planning; broader environments are needed to test how far the conclusions generalize. We also leave retraining-based representation choices, especially VQ macro-actions, to future work. VQ could provide a discrete, data-supported high-level action space by restricting planning to learned codebook entries, but it changes the trained representation rather than only the inference procedure. In this work, the cost of retraining and tuning these variants led us to prioritize inference-time interventions for the trained continuous Hi-LeWM model. Our results should therefore not be read as ruling out VQ-based hierarchies, but as separating that architectural question from the support-constrained planning mechanism studied here.

Interpretation should also be conditioned on model scale.  \citet{zhang2026hierarchical} report substantially stronger PushT performance with a higher-capacity DINO-WM setup, so our findings should not be read as a general limitation of hierarchy. Rather, they identify the conditions under which this hierarchical interface transfers to compact frozen LeWM: subgoals must remain data-supported, executable, and matched to the temporal scale at which they are used.

\clearpage
\bibliographystyle{plainnat}
\bibliography{references}

\clearpage

\appendix

\section{Reproducibility and Code Availability}
\label{app:reproducibility}

We provide the code and checkpoints as a Zenodo artifact at \href{https://doi.org/10.5281/zenodo.21353240}{doi:10.5281/zenodo.21353240}. The implementation builds on the stable-worldmodel library~\citep{maes_lld2026swm}, which provides the environment interfaces, model-predictive-control components, and baseline world-model infrastructure used in our experiments. The remaining appendix sections document the training setup, evaluation protocol, planning configurations, and hyperparameter sweeps needed to reproduce the reported comparisons.

\subsection{Computational Resources}

Training the final Hi-LeWM models required 7.56 hours for PushT on a single NVIDIA A100 GPU and 12.80 hours for Cube on a single NVIDIA H100 GPU. Evaluation was performed using 50 episodes per run. Individual evaluation runs typically completed within tens of minutes, with observed wall-clock times ranging from approximately 5 to 58 minutes depending on the environment and planning budget.

\section{Hi-LeWM Training Details}
\label{app:Hi-LeWM-details}

\subsection{High-Level Architecture}
The trainable high-level branch consists of a latent macro-action encoder $g$ and a high-level predictor $p_{\mathrm{hi}}$. The encoder maps a primitive action chunk between two consecutive waypoints to a fixed-dimensional latent macro-action $\ell_i \in \mathbb{R}^{d_\ell}$. Based on preliminary experiments, we use $d_\ell=32$ for the main Hi-LeWM configuration.

The latent macro-action encoder $g$ is implemented as a transformer encoder over action tokens. Each primitive action is first linearly projected to a model dimension of $192$, a learnable \texttt{[CLS]} token is prepended, and learned positional embeddings are added. The sequence is then processed by a $2$-layer transformer with $4$ attention heads, feed-forward dimension $768$, and dropout $0.1$. The final hidden state of the \texttt{[CLS]} token is projected to $\mathbb{R}^{d_\ell}$ to obtain the macro-action latent. For the main model, the maximum action-chunk length is $15$.

The high-level predictor $p_{\mathrm{hi}}$ operates in the same latent space as the pretrained LeWM encoder. It receives the current waypoint latent together with the corresponding macro-action latent and predicts the next waypoint latent. In our implementation, the predictor uses depth $6$, $16$ attention heads, head dimension $64$, feed-forward dimension $2048$, and dropout $0.1$. Since the LeWM latent dimension is $d_z = 192$, macro-actions are projected from $\mathbb{R}^{d_\ell}$ to $\mathbb{R}^{192}$ before conditioning the predictor.

\begin{figure}[tbp]
    \centering
    \includegraphics[width=0.86\linewidth]{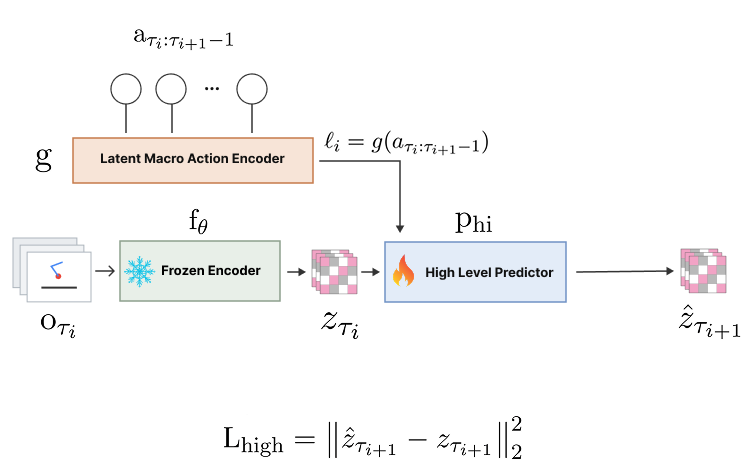}
    \caption{
    High-level training architecture of Hi-LeWM for one waypoint transition. The frozen LeWM encoder maps the current waypoint observation $o_{\tau_i}$ to the latent state $z_{\tau_i}=f_\theta(o_{\tau_i})$. In parallel, the trainable macro-action encoder compresses the primitive action chunk $a_{\tau_i:\tau_{i+1}-1}$ into a latent macro-action $\ell_i=g(a_{\tau_i:\tau_{i+1}-1})$. The high-level predictor $p_{\mathrm{hi}}$ receives $(z_{\tau_i},\ell_i)$ and predicts the next waypoint latent $\hat z_{\tau_{i+1}}$, supervised by a mean-squared latent prediction loss against the encoded target $z_{\tau_{i+1}}$.
    }
    \label{fig:architecture}
\end{figure}

The resulting high-level training path is summarized in \autoref{fig:architecture}, which shows how frozen observation latents and learned macro-action latents are combined to predict the next waypoint latent.

\subsection{Waypoint Sampling Strategy}
High-level supervision is defined by sampling ordered waypoint indices
\[
  \tau_1 < \tau_2 < \cdots < \tau_N.
\]
In the main model, we use a random waypoint strategy (\texttt{random\_sorted}) with $N=5$ waypoints. The first waypoint is anchored at the end of the low-level context window, and the remaining waypoints are sampled in the future subject to a minimum stride of $1$ and a maximum total span of $15$. Consequently, the temporal gaps
\[
  k_i = \tau_{i+1}-\tau_i
\]
are variable, so the associated macro-actions summarize action segments of non-constant duration.

For each consecutive pair $(\tau_i,\tau_{i+1})$, the primitive action chunk
\[
  a_{\tau_i:\tau_{i+1}-1}
\]
is encoded into a latent macro-action $\ell_i$. This sampling procedure defines the temporal abstraction learned by the high-level model: different waypoint gaps correspond to different effective durations of the macro-action.
\subsection{Optimization Details}
High-level training uses the AdamW optimizer with a learning rate of $5 \times 10^{-5}$ and weight decay of $10^{-3}$. We use a batch size of $128$, gradient clipping with a maximum norm of $1.0$, and \texttt{bf16} precision.

Unless otherwise stated, the low-level LeWM encoder, low-level predictor $p_{\mathrm{lo}}$, and low-level action encoder are kept frozen, and only the newly introduced high-level modules $g$ and $p_{\mathrm{hi}}$ are optimized. We also experimented with joint end-to-end training of both the high-level components and the low-level predictor and encoder. However, this did not yield measurable performance gains. We therefore retain the frozen low-level setup, which is also computationally more efficient.

The resulting Hi-LeWM model has 30.5M parameters in total, of which 12.5M are trainable high-level parameters and 18.0M are frozen low-level LeWM parameters.

All models discussed in this paper were trained for up to 15 epochs, except for the VQ variant, which was trained for 50 epochs.

\subsection{High-Level Objective}
Let $z_{\tau_i}$ denote the latent embedding of waypoint $\tau_i$, and let $\ell_i$ denote the macro-action latent extracted from the primitive actions between $\tau_i$ and $\tau_{i+1}$. The high-level predictor is trained to predict the next waypoint latent from the current waypoint latent and the corresponding macro-action latent using a mean-squared latent prediction loss:
\[
\mathcal{L}_{\mathrm{high}}
=
\mathbb{E}\big[
\|p_{\mathrm{hi}}(z_{\tau_i},\ell_i)-z_{\tau_{i+1}}\|_2^2
\big].
\]
For the continuous variants, this is the primary training objective.

\subsection{Alternative High-Level Macro-Action Representation Designs}
\label{ssec:VQ-vae}
Our reference Hi-LeWM configuration uses continuous latent macro-actions, random waypoint sampling, and latent action dimension $d_\ell=32$. This dimensionality was chosen empirically. A larger choice, $d_\ell=192$, was initially considered to match the LeWM latent-state dimension, but in practice it made high-level planning substantially less tractable by enlarging the continuous search space optimized by CEM. We therefore retain the more compact setting $d_\ell=32$ in the main model.

\subsubsection{Fixed-Stride Variant.} We also evaluate a fixed-stride variant of the waypoint construction. Whereas the reference model samples waypoints randomly within a bounded span, the fixed-stride variant places them deterministically every $5$ steps. To preserve the overall temporal extent of the supervision signal, we use $N=4$ waypoints, so that the total span remains $(N-1)\times 5 = 15$, matching the maximum span of the reference configuration. Within this setting, we report two continuous latent action dimensions, $d_\ell=32$ and $d_\ell=8$.

\subsubsection{VQ Macro-Action Encoder.}
We also evaluate a vector-quantized variant of the macro-action encoder, following the discrete bottleneck idea introduced by VQ-VAE~\citep{oord2018neuraldiscreterepresentationlearning}. Unlike an image VQ-VAE, this module quantizes only action chunks: a transformer encoder maps each primitive action segment to a continuous macro-action latent, which is then replaced by its nearest entry in a learned codebook. The quantized latent is passed to the high-level predictor, so high-level transitions are trained on discrete macro-action representations rather than arbitrary continuous vectors.

\paragraph{Training Objective.}
The VQ variant augments the high-level prediction objective with the standard auxiliary terms used for vector quantization:
\[
  \mathcal{L}_{\mathrm{VQ}}
  =
  \mathcal{L}_{\mathrm{high}}
  +
  \lambda_{\mathrm{rec}} \mathcal{L}_{\mathrm{rec}}
  +
  \lambda_{\mathrm{commit}} \mathcal{L}_{\mathrm{commit}}
  +
  \lambda_{\mathrm{code}} \mathcal{L}_{\mathrm{code}} .
\]
Here, $\mathcal{L}_{\mathrm{high}}$ is the waypoint-level latent prediction loss. The reconstruction term $\mathcal{L}_{\mathrm{rec}}$ trains a decoder to reconstruct the original primitive action chunk from the quantized macro-action latent. The commitment loss $\mathcal{L}_{\mathrm{commit}}$ encourages the encoder output to remain close to the selected code, while the codebook loss $\mathcal{L}_{\mathrm{code}}$ updates the selected code toward the encoder output. We use the straight-through quantized latent for high-level prediction, so gradients from $\mathcal{L}_{\mathrm{high}}$ still train the encoder while the forward pass remains codebook-constrained.

\paragraph{Role of the Decoder.}
The decoder is used only during training. It is not part of the acting policy and is not used to generate primitive actions at evaluation time. Its purpose is to regularize the discrete bottleneck by forcing each code to retain information about the action segment it represents. Thus, the VQ module acts as a macro-action representation learner rather than as a separate low-level controller.

\paragraph{Configurations.}
We report two VQ configurations with codebook sizes $16$ and $128$. These variants test whether restricting the high-level planner to a discrete macro-action vocabulary improves support and planning stability compared with continuous macro-action latents.

Taken together, these variants compare three high-level design choices: variable- versus fixed-duration waypoint transitions, wider versus narrower latent macro-action spaces, and continuous versus quantized macro-action representations. The resulting PushT performance is summarized in Table~\ref{tab:Hi-LeWM-representation-results}.

\begin{table*}[h]
  \centering
  \renewcommand{\arraystretch}{1.12}
  \setlength{\tabcolsep}{7pt}
  \small
  \begin{tabular}{lllcc}
    \toprule
    \textbf{Sampling} & \textbf{Macro-action} & \textbf{Representation size} &
    \begin{tabular}[c]{@{}c@{}}\textbf{$d=50$}\\ \textbf{Success rate}
    \end{tabular} &
    \begin{tabular}[c]{@{}c@{}}\textbf{$d=75$}\\ \textbf{Success rate}
    \end{tabular} \\
    \midrule
    \rowcolor{blue!8}
    random & Continuous & $d_\ell=32$ & \textbf{46} & 20 \\
    fixed stride & Continuous & $d_\ell=32$ & 42 & 20 \\
    fixed stride & Continuous & $d_\ell=8$  & 42 & 20 \\
    random & VQ & 16 codes & 38 & 20 \\
    random & VQ & 128 codes & 44 & \textbf{34} \\
    \bottomrule
  \end{tabular}
  \caption{Best PushT success rate (\%) for each high-level representation design. The highlighted first row corresponds to the reference Hi-LeWM model used throughout the paper.}
  \label{tab:Hi-LeWM-representation-results}
\end{table*}

The representation sweep shows that no macro-action parameterization is uniformly best. At $d=50$, the reference continuous model with random waypoint sampling and $d_\ell=32$ performs best, reaching $46\%$ success. Reducing the continuous macro-action dimension to $d_\ell=8$ does not improve performance, and using fixed-stride waypoints also slightly reduces success. At $d=75$, however, the VQ-128 variant performs best, improving from $20\%$ to $34\%$. This suggests that a discrete macro-action bottleneck can help in the longest-horizon setting by restricting high-level search to a smaller set of action abstractions. In contrast, the smaller VQ-16 codebook does not improve over the continuous variants, indicating that overly aggressive discretization removes useful macro-action expressivity.

Despite the promising results, we do not use VQ-128 as the main Hi-LeWM variant. The discrete representation introduces additional architectural and optimization complexity, including codebook learning, action-chunk reconstruction, commitment and codebook losses, and a separate discrete planning interface. Since the goal of this work is to evaluate whether a minimal hierarchy can improve LeWM without substantially changing its model class or planning setup, we retain the continuous $d_\ell=32$ model as the reference architecture. Overall, these results suggest that constraining the high-level action space can improve robustness at longer horizons, but doing so introduces an architectural burden that goes beyond the minimal extension studied in the main paper.

\FloatBarrier
\section{Evaluation Configurations}
\label{app:eval-configs}

\subsection{Cube Results and Scope}
\label{app:cube-results}

We report Cube results in the appendix because \texttt{OGBCube-v0} is less diagnostic for the long-horizon planning failure that motivates the main text. Unlike PushT, the flat LeWM baseline remains comparatively stable as the goal offset increases: it obtains $65.3\%$, $52.0\%$, and $53.3\%$ at $d\in\{25,50,75\}$, respectively. This makes Cube useful as an additional robustness check, but less suited for isolating the regime in which flat latent CEM degrades and hierarchy could provide a clear long-horizon advantage.

The full Cube sweep is included in Table~\ref{tab:combined-final-table}. Hi-LeWM and Hi-LeWM-C are competitive on Cube, and the constrained variant reaches $68.0\%$, $54.7\%$, and $69.3\%$ at $d\in\{25,50,75\}$. These results are consistent with the support-constrained planning interpretation, but they do not drive the main claim because Cube does not exhibit the same monotonic horizon-induced collapse as PushT.

\subsection{Evaluation Protocol}
\label{app:eval-protocol}

We follow the evaluation methodology of the original LeWorldModel paper~\citep{maes2026leworldmodel} so that our reproduced LeWM baseline and all Hi-LeWM variants are compared under the same episode-construction and goal-definition procedure. The evaluation code uses the stable-worldmodel interfaces for the environments, MPC solvers, and baseline components~\citep{maes_lld2026swm}.

Each evaluation episode is built from an expert demonstration trajectory. We first sample a start observation from the trajectory. The goal is then defined as a future observation from that same trajectory; in our benchmark setting, this corresponds to the demonstrated end configuration of the expert segment. The agent therefore receives the current observation together with a goal observation and must act so as to reach that future target state.

To vary task difficulty, we control the temporal distance between the sampled start and goal using the offset
$
d \in \{25, 50, 75\},
$
where $d$ denotes the number of environment steps separating the start observation from the goal observation. Larger values of $d$ correspond to longer-horizon planning problems.

For each planning configuration, we evaluate the agent on a fixed set of sampled episodes and report the success rate, i.e., the percentage of episodes in which the agent reaches the goal. In the main evaluation protocol, each configuration is evaluated on 50 episodes. Unless otherwise noted, reported values are averaged over three random seeds.

\subsection{LeWM Baseline Configurations}
\label{app:lewm-baseline-configs}

Table~\ref{tab:orig-lewm-baselines-appendix} reports the planning configurations selected for the LeWorldModel baseline on PushT and Cube. For each environment and goal offset, we sweep the flat planning hyperparameters and retain the best-performing setting. These results provide the reference point used in the main paper when comparing flat LeWM against Hi-LeWM.

\begin{table}[h]
  \centering
  \small
  \setlength{\tabcolsep}{5.4pt}
  \renewcommand{\arraystretch}{1.12}

  \begin{minipage}[t]{0.48\textwidth}
    \centering
    \begin{tabular}{llccc}
      \toprule
      \multicolumn{5}{c}{\textbf{PushT}} \\
      \cmidrule(lr){1-5}
      Goal & Budget & $h$ & $r$ & Success (\%) \\
      \midrule
      \textbf{$d{=}25$}
      & \cellcolor{bestflat}50 & \cellcolor{bestflat}5 & \cellcolor{bestflat}5 & \cellcolor{bestflat}\textbf{94.0 $\pm$ 2.0} \\
      \midrule
      \multirow{3}{*}{\textbf{$d{=}50$}}
      & 100 & 5  & 1 & 36.0 $\pm$ 7.2 \\
      & \cellcolor{bestflat}100 & \cellcolor{bestflat}5 & \cellcolor{bestflat}5 & \cellcolor{bestflat}\textbf{52.7 $\pm$ 5.0} \\
      & 100 & 10 & 5 & 22.0 $\pm$ 8.0 \\
      \midrule
      \multirow{3}{*}{\textbf{$d{=}75$}}
      & \cellcolor{bestflat}150 & \cellcolor{bestflat}5 & \cellcolor{bestflat}5 & \cellcolor{bestflat}\textbf{18.0 $\pm$ 2.0} \\
      & 150 & 10 & 5 & 10.7 $\pm$ 1.2 \\
      & 150 & 15 & 5 & 3.3 $\pm$ 2.3 \\
      \bottomrule
    \end{tabular}
  \end{minipage}
  \hfill
  \begin{minipage}[t]{0.48\textwidth}
    \centering
    \begin{tabular}{llccc}
      \toprule
      \multicolumn{5}{c}{\textbf{Cube}} \\
      \cmidrule(lr){1-5}
      Goal & Budget & $h$ & $r$ & Success (\%) \\
      \midrule
      \textbf{$d{=}25$}
      & \cellcolor{bestflat}50 & \cellcolor{bestflat}5 & \cellcolor{bestflat}5 & \cellcolor{bestflat}\textbf{65.3 $\pm$ 4.2} \\
      \midrule
      \multirow{3}{*}{\textbf{$d{=}50$}}
      & 100 & 5  & 1 & 48.0 $\pm$ 2.0 \\
      & \cellcolor{bestflat}100 & \cellcolor{bestflat}5 & \cellcolor{bestflat}5 & \cellcolor{bestflat}\textbf{52.0 $\pm$ 3.5} \\
      & 100 & 10 & 5 & 45.3 $\pm$ 1.2 \\
      \midrule
      \multirow{3}{*}{\textbf{$d{=}75$}}
      & \cellcolor{bestflat}150 & \cellcolor{bestflat}5 & \cellcolor{bestflat}5 & \cellcolor{bestflat}\textbf{53.3 $\pm$ 7.0} \\
      & 150 & 10 & 5 & 40.7 $\pm$ 8.3 \\
      & 150 & 15 & 5 & 41.3 $\pm$ 4.2 \\
      \bottomrule
    \end{tabular}
  \end{minipage}

  \caption{Original LeWorldModel baseline results on PushT and Cube. We report the planning horizon $h$, receding-horizon execution length $r$, and success rate. The action block size is fixed to 5 and the CEM budget is fixed to $300 \times 30$ for all runs. Results are mean $\pm$ standard deviation over three seeds. Shaded cells mark the best configuration within each environment and goal distance.}
  \label{tab:orig-lewm-baselines-appendix}
\end{table}

\begin{table}[H]
  \centering
  \scriptsize
  \setlength{\tabcolsep}{3.2pt}
  \renewcommand{\arraystretch}{1.08}

  \begin{tabular}{@{}lccccc c !{\color{black!35}\vrule width 0.45pt} cccc@{}}
    \toprule
    \textbf{Goal}
    & \textbf{Budget}
    & $\boldsymbol{h_h}$
    & $\boldsymbol{h_{\mathrm{replan}}}$
    & $\boldsymbol{h_\ell}$
    & $\boldsymbol{r_\ell}$
    & \textbf{Staged}
    & \textbf{PushT Hi-LeWM}
    & \textbf{PushT Hi-LeWM-C}
    & \textbf{Cube Hi-LeWM}
    & \textbf{Cube Hi-LeWM-C} \\
    \midrule

    \multicolumn{11}{@{}l}{\textbf{$d=25$}} \\
    & 50 & 1 & 5 & 2 & 1 &  & \meanstd{89.3}{4.1} & \textbf{\meanstd{88.7}{4.1}} & \meanstd{67.3}{1.9} & \meanstd{67.3}{3.4} \\
    & 50 & 2 & 5 & 2 & 1 &  & \meanstd{28.0}{8.5} & \meanstd{28.7}{5.2} & \meanstd{69.3}{2.5} & \meanstd{54.7}{0.9} \\
    & 50 & 1 & 5 & 3 & 1 &  & \meanstd{70.7}{4.1} & \meanstd{75.3}{0.9} & \meanstd{64.0}{0.0} & \meanstd{64.7}{0.9} \\
    & 50 & 1 & 5 & 5 & 1 &  & \meanstd{44.0}{1.6} & \meanstd{40.7}{0.9} & \meanstd{64.0}{3.3} & \meanstd{61.3}{0.9} \\
    & 50 & 1 & 5 & 5 & 5 &  & \meanstd{79.3}{4.7} & \meanstd{80.7}{3.4} & \meanstd{66.0}{2.8} & \textbf{\meanstd{68.0}{4.3}} \\
    & 50 & 2 & 5 & 2 & 2 &  & \meanstd{23.3}{3.8} & \meanstd{24.0}{8.6} & \textbf{\meanstd{70.7}{0.9}} & \meanstd{52.7}{2.5} \\
    \rowcolor{stagedrow}
    & 50 & 2 & 5 & 2 & 1 & $\checkmark$ & \textbf{\meanstd{90.7}{6.2}} & \meanstd{70.7}{5.2} & \meanstd{65.3}{2.5} & \meanstd{60.7}{2.5} \\

    \midrule
    \multicolumn{11}{@{}l}{\textbf{$d=50$}} \\
    & 100 & 1 & 5 & 2 & 1 &  & \meanstd{38.7}{9.0} & \meanstd{48.7}{3.8} & \textbf{\meanstd{52.0}{3.3}} & \textbf{\meanstd{54.7}{3.4}} \\
    & 100 & 2 & 5 & 2 & 2 &  & \meanstd{27.3}{1.9} & \meanstd{32.0}{2.8} & \textbf{\meanstd{52.0}{0.0}} & \meanstd{44.0}{2.8} \\
    & 100 & 2 & 5 & 2 & 1 &  & \meanstd{35.3}{6.6} & \meanstd{34.7}{3.4} & \meanstd{50.7}{3.4} & \meanstd{48.7}{2.5} \\
    & 100 & 2 & 5 & 5 & 1 &  & \meanstd{16.0}{5.9} & \meanstd{22.0}{5.9} & \meanstd{48.0}{1.6} & \meanstd{42.7}{2.5} \\
    & 100 & 2 & 5 & 5 & 5 &  & \meanstd{20.7}{5.7} & \meanstd{18.7}{4.1} & \meanstd{48.0}{1.6} & \meanstd{44.0}{1.6} \\
    \rowcolor{stagedrow}
    & 100 & 2 & 5 & 2 & 1 & $\checkmark$ & \textbf{\meanstd{42.0}{2.8}} & \textbf{\meanstd{64.0}{3.3}} & \meanstd{46.7}{2.5} & \meanstd{48.0}{0.0} \\
    & 100 & 2 & 3 & 2 & 1 &  & \meanstd{34.0}{4.3} & \meanstd{38.7}{6.2} & \textbf{\meanstd{52.0}{1.6}} & \meanstd{48.0}{2.8} \\
    & 100 & 3 & 5 & 2 & 1 &  & \meanstd{24.0}{4.3} & \meanstd{20.7}{5.7} & \meanstd{47.3}{0.9} & \meanstd{29.3}{5.2} \\

    \midrule
    \multicolumn{11}{@{}l}{\textbf{$d=75$}} \\
    & 150 & 1 & 5 & 2 & 1 &  & \meanstd{12.7}{3.8} & \meanstd{19.3}{0.9} & \textbf{\meanstd{66.7}{6.2}} & \textbf{\meanstd{69.3}{6.8}} \\
    & 150 & 2 & 5 & 2 & 2 &  & \meanstd{13.3}{5.0} & \meanstd{18.7}{4.1} & \meanstd{56.7}{4.1} & \meanstd{43.3}{5.2} \\
    & 150 & 2 & 5 & 2 & 1 &  & \textbf{\meanstd{15.3}{3.8}} & \textbf{\meanstd{32.7}{3.8}} & \meanstd{58.7}{4.7} & \meanstd{42.7}{4.1} \\
    & 150 & 2 & 5 & 5 & 1 &  & \meanstd{10.0}{4.3} & \meanstd{16.0}{3.3} & \meanstd{42.0}{1.6} & \meanstd{37.3}{2.5} \\
    & 150 & 2 & 5 & 5 & 5 &  & \meanstd{10.7}{4.1} & \meanstd{17.3}{0.9} & \meanstd{46.7}{0.9} & \meanstd{42.0}{5.9} \\
    \rowcolor{stagedrow}
    & 150 & 2 & 5 & 2 & 1 & $\checkmark$ & \textbf{\meanstd{15.3}{4.1}} & \meanstd{22.0}{3.3} & \meanstd{49.3}{0.9} & \meanstd{50.0}{4.9} \\
    & 150 & 2 & 3 & 2 & 1 &  & \meanstd{13.3}{0.9} & \meanstd{25.3}{8.2} & \meanstd{58.7}{4.7} & \meanstd{41.3}{2.5} \\
    & 150 & 3 & 5 & 2 & 1 &  & \meanstd{14.0}{0.0} & \meanstd{12.7}{2.5} & \meanstd{46.0}{5.9} & \meanstd{40.0}{3.3} \\

    \bottomrule
  \end{tabular}
  \caption{Planner success rates across goal offsets. Each row is one Hi-LeWM planning configuration. Results report mean $\pm$ standard deviation across seeds. \textsc{Hi-LeWM} denotes standard high-level CEM, while \textsc{Hi-LeWM-C} denotes constrained empirical-macro CEM. Bold marks the best result within each benchmark, goal offset, and model variant. Green marks staged high-level execution.}
  \label{tab:combined-final-table}

\end{table}

\clearpage
\subsection{Hi-LeWM Planning Sweep}
\label{app:Hi-LeWM-planning-configs}

Table~\ref{tab:combined-final-table} reports the main planning sweep for Hi-LeWM on PushT and Cube. We vary the high-level horizon, the high-level replanning interval, the low-level planning horizon, the low-level receding-horizon execution length, and whether the controller executes staged subgoals. For each goal offset, we highlight the best online configuration and the best staged configuration.

\subsection{Hi-LeWM CEM Hyperparameters}
\label{app:Hi-LeWM-cem}

In addition to the horizon choices above, Table~\ref{tab:Hi-LeWM-cem-configs} reports the CEM sampling budgets used for Hi-LeWM. These settings are shared across PushT and Cube for the hierarchical evaluations.

\begin{table}[H]
  \centering
  \small
  \setlength{\tabcolsep}{5pt}
  \renewcommand{\arraystretch}{1.10}
  \begin{tabular}{@{}lccccccc@{}}
    \toprule
    \textbf{Goal}
    & \textbf{Budget}
    & \multicolumn{3}{c}{\textbf{High-level CEM}}
    & \multicolumn{3}{c}{\textbf{Low-level CEM}} \\
    \cmidrule(lr){3-5}
    \cmidrule(lr){6-8}
    &
    & \textbf{Samples} & \textbf{Steps} & \textbf{Top-$k$}
    & \textbf{Samples} & \textbf{Steps} & \textbf{Top-$k$} \\
    \midrule
    $d=25$ & 50  & 900  & 20 & 10 & 300  & 30 & 150 \\
    $d=50$ & 100 & 1500 & 40 & 10 & 900  & 30 & 150 \\
    $d=75$ & 150 & 1200 & 60 & 10 & 1200 & 30 & 150 \\
    \bottomrule
  \end{tabular}
  \caption{CEM planning hyperparameters used for Hi-LeWM evaluations for both PushT and Cube environments.}
  \label{tab:Hi-LeWM-cem-configs}
\end{table}

\subsection{Empirical-Macro CEM Details}
\label{app:empirical_macro_cem}

Standard high-level CEM samples full latent macro-action sequences over the high-level horizon directly from a continuous latent distribution. Empirical-macro CEM instead constructs an empirical bank $\mathcal{B} = \{\xi_j\}_{j=1}^{N}$, where each $\xi_j$ is a latent macro-action sequence induced by real training behavior. The bank is built by sampling contiguous action spans from the training action stream, grouping them into macro-action chunks, and encoding them with the learned macro-action encoder.

At planning time, candidates are generated as
\[
    \xi = \xi_{\mathrm{bank}} + \lambda_{\mathrm{res}} \epsilon,
    \qquad
    \xi_{\mathrm{bank}} \sim \mathrm{Unif}(\mathcal{B}),
    \qquad
    \epsilon \sim \mathcal{N}(\mu, \Sigma).
\]
The residual distribution is initialized with zero mean and diagonal standard deviation controlled by the residual-scale hyperparameter $\lambda_{\mathrm{res}}$. CEM evaluates the full candidates, selects elites according to the model-based planning cost, and refits only the residual distribution from the elite residuals:
\[
    \epsilon_{\mathrm{elite}}
    =
    \xi_{\mathrm{elite}} - \xi_{\mathrm{bank}, \mathrm{elite}}.
\]

The empirical component is not optimized directly. Each CEM iteration resamples anchors from the empirical bank and learns how to locally perturb them. One zero-residual candidate is included for the sampled anchor in each iteration, so the planner evaluates an unmodified empirical macro-action sequence from the current sampled batch.

For staged planning, we compare sequence-level bank sampling and independent stage sampling. Sequence sampling preserves temporal correlations between successive macro-actions, while independent sampling only preserves marginal per-stage support. The strongest staged results used sequence sampling.

\section{Diagnostic Details}
\label{app:diagnostic_details}

\subsection{Latent-to-Pixel Decoder Probe}
\label{app:decoder_probe}

We train a lightweight decoder probe to test whether the latent states used by the hierarchical world model retain visually meaningful information. The probe is not used during policy execution. It is used only for diagnostics: if both encoded future latents and predicted future latents can be decoded into recognizable observations, then failures in control are less likely to be caused by a complete loss of visual information in the latent representation.

For a high-level waypoint transition, $z_{\tau_i}=f_\theta(o_{\tau_i})$ denotes the current waypoint latent and $z_{\tau_{i+1}}=f_\theta(o_{\tau_{i+1}})$ denotes the encoded future waypoint latent. The model predicts
$
    \hat z_{\tau_{i+1}}=p_{\mathrm{hi}}(z_{\tau_i},\ell_i),
$
where $\ell_i$ is the macro-action latent for the corresponding action chunk. We train a separate decoder $D_\phi$ to reconstruct waypoint images from latents,
\[
    \tilde o = D_\phi(z), \qquad
    D_\phi:\mathbb{R}^{d_z}\rightarrow\mathbb{R}^{3\times224\times224}.
\]
All world-model parameters are frozen; only the decoder parameters are updated.

The decoder is a query-based transformer that maps a single latent vector to a $14\times14$ grid of image patches with patch size $16$. Each block performs self-attention over learned patch queries and cross-attention to the latent token, followed by an MLP. The predicted patches are then reassembled into a $224\times224$ RGB image.

The probe is trained on sampled future waypoints from the task dataset. We first train it on true encoded waypoint latents,
\[
    \mathcal{L}_{\mathrm{true}}
    =
    \left\|D_\phi(z_{\tau_{i+1}}) - o_{\tau_{i+1}}\right\|_2^2 ,
\]
and then expose it to predicted high-level latents using
\[
    \mathcal{L}_{\mathrm{probe}}
    =
    \left\|D_\phi(z_{\tau_{i+1}}) - o_{\tau_{i+1}}\right\|_2^2
    +
    \lambda_{\mathrm{pred}}
    \left\|D_\phi(\hat z_{\tau_{i+1}}) - o_{\tau_{i+1}}\right\|_2^2 ,
\]
with $\lambda_{\mathrm{pred}}=0.5$. We optimize with AdamW, learning rate $2\times10^{-4}$, weight decay $10^{-4}$, batch size 128, and a 90/10 train/validation split.

\begin{figure}[h]
  \centering
  \includegraphics[width=\linewidth]{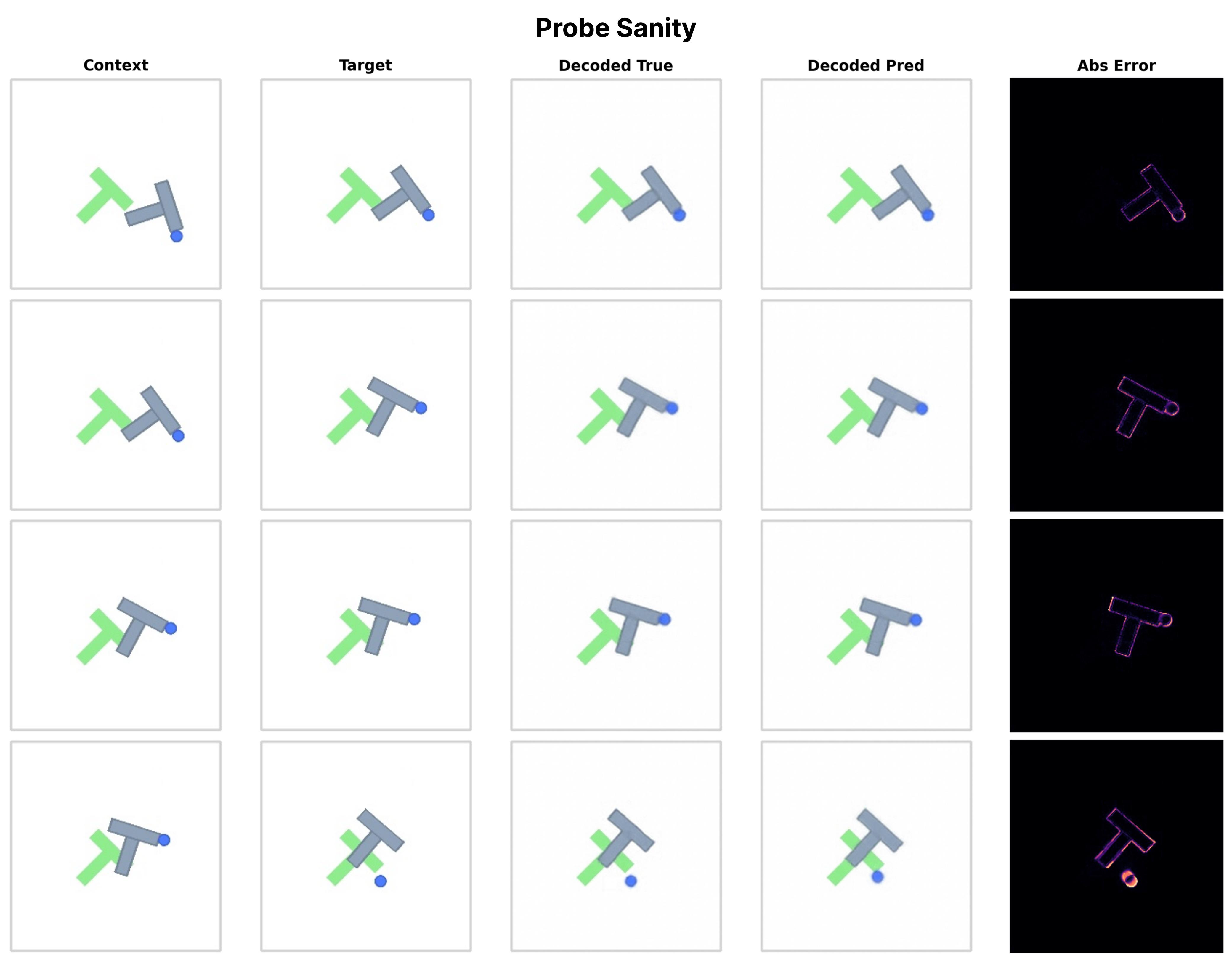}
  \caption{
    Decoder probe sanity check. Each row shows one validation example. Columns show the context observation, true future observation, reconstruction from the true latent, reconstruction from the predicted latent, and absolute pixel error. The reconstructions preserve the main geometry and agent position, suggesting that both true and predicted waypoint latents remain visually decodable.
  }
  \label{fig:decoder_probe_sanity}
\end{figure}

Figure~\ref{fig:decoder_probe_sanity} shows this qualitatively, and the validation metrics over 8 batches support the same conclusion. Reconstructions from true latents achieve pixel mean squared error (MSE) $0.0040 \pm 0.00007$, peak signal-to-noise ratio (PSNR) $36.89 \pm 0.07$ dB, and structural similarity index measure (SSIM) $0.9913 \pm 0.0001$. Reconstructions from predicted latents are worse, but still strong: pixel mean squared error (MSE) $0.0097 \pm 0.0004$, peak signal-to-noise ratio (PSNR) $33.08 \pm 0.19$ dB, and structural similarity index measure (SSIM) $0.9834 \pm 0.0006$. So the probe decodes true latents very accurately, and even predicted latents remain highly decodable, with only a moderate drop in fidelity.

The probe therefore provides a useful sanity check for the latent space. Since predicted latents still decode into plausible observations with high SSIM, the main control failures are more likely due to high-level planning, temporal alignment, or subgoal selection than to a completely non-informative latent representation.

\end{document}